\documentclass[runningheads]{llncs}

\usepackage{eccv}

\usepackage{eccvabbrv}

\usepackage{graphicx}
\usepackage{booktabs}
\usepackage{algorithm,algpseudocode}

\usepackage{makecell}
\usepackage{color}
\usepackage{enumitem}
\usepackage{multirow}
\usepackage{bbm}
\usepackage{arydshln}
\usepackage{subcaption}
\usepackage{bm}
\usepackage[accsupp]{axessibility}  % Improves PDF readability for those with disabilities.

\newcommand{\bhr}[1]{\textcolor{black}{#1}}

\newcommand{\ms}[1]{\textcolor{magenta}{#1}}

\newcommand{\black}[1]{\textbf{\textcolor{black}{#1}}}
\newcommand{\bred}[1]{\textbf{\textcolor{red}{#1}}}

\newcommand{\bb}[1]{\mathbf{#1}}
\newcommand{\bx}{\bb{x}}
\newcommand{\by}{\bb{y}}
\newcommand{\bxsrc}{\bb{x}^{\text{src}}}
\newcommand{\psrc}{p^{\text{src}}}
\newcommand{\ptgt}{p^{\text{tgt}}}

\newcommand{\bysrc}{\bb{y}^{\text{src}}}
\newcommand{\bxtgt}{\bb{x}^{\text{tgt}}}
\newcommand{\hbxtgt}{\bb{\hat{x}}^{\text{tgt}}}
\newcommand{\bytgt}{\bb{y}^{\text{tgt}}}
\newcommand{\at}{\alpha_t}
\newcommand{\atpone}{\alpha_{t+1}}
\newcommand{\atmone}{\alpha_{t-1}}
\newcommand{\fth}{f_\theta}
\newcommand{\epth}{\epsilon_\theta}
\newcommand{\hepth}{\hat{\epsilon}_\theta}

\newcommand{\Msrc}{\bb{M}^\text{src}}
\newcommand{\Mtgt}{\bb{M}^\text{tgt}}

\usepackage{hyperref}

\usepackage{orcidlink}

\begin{document}

%\title{Noise Correction via Prompt Interpolation for Diffusion-Based Image-to-Image Translation} 
\title{Diffusion-Based Image-to-Image Translation by Noise Correction via Prompt Interpolation} 

\titlerunning{Diffusion-Based Image-to-Image Translation by Noise Correction}

% Include the authors' OCRID for the camera-ready version, spif at all possible.

\author{
Junsung Lee\inst{1} \qquad\quad
Minsoo Kang\inst{1}\qquad\quad
Bohyung Han\inst{1,2}
}

\authorrunning{J. Lee et al.}

\institute{
\textsuperscript{1}ECE \& \textsuperscript{2}IPAI, Seoul National University\\
\email{\{leejs0525,kminsoo,bhhan\}@snu.ac.kr}}

\maketitle

% !TEX root = ./../main.tex
\begin{abstract}
	We propose a simple but effective training-free approach tailored to diffusion-based image-to-image translation. 
	Our approach revises the original noise prediction network of a pretrained diffusion model by introducing a noise correction term.
	%, which is based on the progressive interpolation of the textual prompts corresponding to a given source image and a desired target image.
	% Specifically, we suggest original reverse process tailored to image-to-image translation by replacing denoising network prediction with a new model based on the progressive interpolation between source and target prompts.
	%Specifically, we suggest modified DDIM reverse process tailored to image-to-image translation by replacing noise prediction with progressive and adaptive interpolation between source and target prompt embeddings.
	% In our proposed process, we utilize the reconstruction process for the source image to preserve the structure or background of the source image, and introduce an additional noise correction term to carefully edit only the region of interest simultaneously. 
	%Compared to previous methods, we introduce an additional noise correction term that can be applied into the noise calculated in reverse process of source image.
	We formulate the noise correction term as the difference between two noise predictions; one is computed from the denoising network with a progressive interpolation of the source and target prompt embeddings, while the other is the noise prediction with the source prompt embedding.
	The final noise prediction network is given by a linear combination of the standard denoising term and the noise correction term, where the former is designed to reconstruct must-be-preserved regions while the latter aims to effectively edit regions of interest relevant to the target prompt.
	Our approach can be easily incorporated into existing image-to-image translation methods based on diffusion models.
	%We experimentally found our proposed noise correction only edits the region of interest in source images.
	Extensive experiments verify that the proposed technique achieves outstanding performance with low latency and consistently improves existing frameworks when combined with them.
	\keywords{training-free image-to-image translation \and diffusion models \and generative modeling} % TODO: need to modify abstract into 150 words
\end{abstract} % FINISH
% !TEX root = ./../main.tex

\section{Introduction}
\label{sec:intro}
%Diffusion probabilistic models have recently drawn significant attention in various computer vision problems including image classification~\cite{li2023your, shipard2023boosting}, segmentation~\cite{tian2023diffuse, xu2023open, li2023open}, image inpainting~\cite{xie2023smartbrush}, and image generation~\cite{rombach2022high, dalle2, saharia2022photorealistic}.
%various domains such as computer vision~\cite{}, natural language processing~\cite{}, and speech processing~\cite{} due to its superior generation quality and diversity.
%Although the generated image is meaningfully aligned with the corresponding text prompt, the application of the text-to-image diffusion models to text-driven image-to-image translation tasks is not straightforward.
The diffusion probabilistic model~\cite{song2021score, ho2020denoising, sohl2015deep, song2021denoising} is currently a dominant framework for image generation. 
It has often been trained to generate high-fidelity images from text prompts~\cite{aditya2022dalle2, rombach2022high, saharia2022photorealistic}, and has also been applied to image-to-image translation given a target text prompt~\cite{kim2022diffusionclip, brooks2023instructpix2pix, kawar2023imagic, lee2023conditional, meng2022sdedit, hertz2023prompt, tumanyan2023plug, parmar2023zero}, where the goal is to modify local regions in a source image based on the target prompt while preserving its background or structure of the image.
%as well as prompt-irrelevant parts.
%Existing text-to-image generation algorithms~\cite{aditya2022dalle2, rombach2022high, saharia2022photorealistic} adopt the diffusion model to generate high-fidelity images from text prompts.
%Another approaches~\cite{kim2022diffusionclip, brooks2023instructpix2pix, kawar2023imagic, lee2023conditional, meng2022sdedit, su2023dual, hertz2023prompt, tumanyan2023plug, parmar2023zero} employ the probabilistic model to perform image-to-image translation, where the goal is to modify local regions in a source image based on a target prompt while preserving its background. 
%
%Although diffusion models are widely used to solve such tasks, it is not straightforward to apply the text-to-image diffusion models to text-driven image-to-image translation, which aims to modify local regions in a source image based on a target prompt while preserving its background.
However, the text-driven image-to-image translation task is an inherently challenging problem, mainly because it is infeasible to find a desirable starting point of the reverse diffusion process for denoising and is difficult to exclusively edit specific regions of generated images without distorting the remaining parts.
%
%is not straightforward.
%Since the number  xtract the final latent using the source image and source prompt, and then simply synthesize the target image based on the latent and the target prompt, which often fails to maintain the background of the input image, leading to poor results.
%For instance, we extract the final latent representation using the source image and prompt, then attempt to synthesize the target image based on this latent representation and the target prompt. This approach often fails to preserve the background of the input image, leading to suboptimal outcomes

%difficult.
%often fails to maintain the background of the input images, leading to poor results. 
%For the image-to-image translation, the diffusion models aim to modify the part region in the source image, defined by the source domain, according to the target domain while preserving the background.
%the simple application of the text-to-image diffusion models achieves poor performance, where we aim to modify the part region in the source image, defined by the source domain, according to the target domain while preserving the background.
%the application of the text-to-image diffusion models to text-driven image-to-image translation tasks often fails to maintain the background of the input images, leading to poor results. 

To tackle the critical challenges, several approaches rely on fine-tuning~\cite{kim2022diffusionclip, brooks2023instructpix2pix, kawar2023imagic} for customizing pretrained diffusion-based denoising networks; they encourage the translated images to reflect the target prompt and preserve the background or the structure in the source image. 
On the other hand, training-free techniques~\cite{lee2023conditional, meng2022sdedit, hertz2023prompt, tumanyan2023plug, parmar2023zero} focus on manipulating denoising strategies used in the reverse process of diffusion models without incurring heavy training costs.
%In this work, we are interested in developing a training-free algorithm for text-driven image-to-image translation.

\begin{figure*}[t!]
	\centering
	\includegraphics[width=1.0\linewidth]{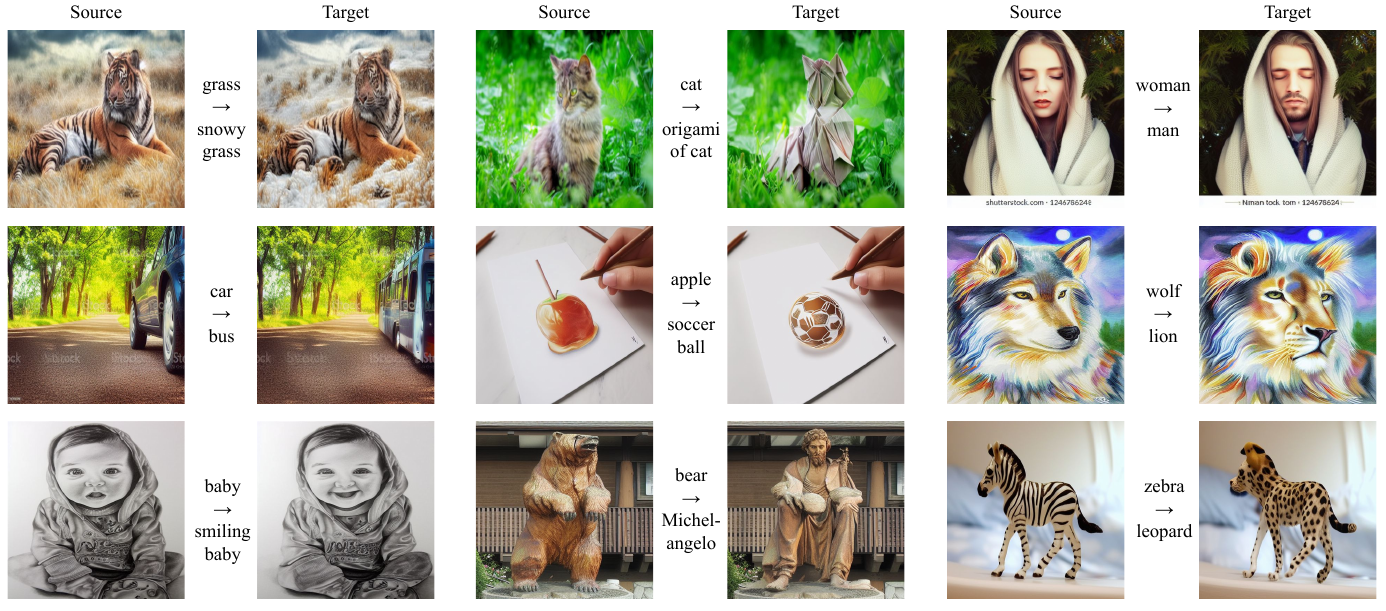}
	\caption{Image-to-image translation results using the proposed method on data sampled from the LAION-5B dataset~\cite{schuhmann2022laion}. Our approach effectively preserves the structure and the background in source images while successfully editing the local region of interest.}
	%Qualitative comparisons between the proposed method and state-of-the-art methods~\cite{hertz2023prompt, tumanyan2023plug, parmar2023zero} on real images sampled from the LAION-5B dataset~\cite{schuhmann2022laion} using the pretrained Stable Diffusion~\cite{rombach2022high}.}
%Qualitative results of Pix2Pix on the Cityscapes dataset.
% ``Output" denotes the ground-truth and ``Original" represents the uncompressed generator.}
\label{fig:visualization_ours}
\end{figure*}

% To this end, we propose a simple but effective method to revise the original noise prediction network tailored to text-driven image-to-image translation by gradually changing the text prompt from the source prompt to the target prompt during the reverse process.
We present a simple but effective training-free image-to-image translation technique, which proposes a variation of the DDIM reverse process.
Our approach estimates the noise correction term to generate desirable images relevant to target prompts, which is achieved by progressive prompt interpolation during the reverse process of diffusion models.
The proposed noise prediction network for image-to-image translation is composed of two parts: (a) the denoising network output given the source latent and the source prompt and (b) a noise correction term defined as the difference between the two noise predictions of the target latent conditioned on the progressively interpolated embeddings and the source text embeddings. 
%to change the direction of convergence towards the target domain
The first term ensures that the target image preserves overall structure and background in the source image while the second term facilitates alignment with the target domain by selectively editing the regions of interest. 
We visualize text-driven image-to-image translation results in \cref{fig:visualization_ours}, which demonstrates the outstanding performance of the proposed approach across various tasks.
%In order to selectively edit the region of interest, a noise correction term is also incorporated into the denoising output using the source latent and source prompt as inputs. 
%The proposed noise correction is efficiently computed by the difference between two noise predictions: the initial one is estimated using the target latent and a smoothly interpolated embeddings between the source and target prompt embeddings while the second one is derived from the target latent with the source prompt embedding. 
%,
%

%
%
%
%
The main contributions of our work are summarized below: 
\begin{itemize}
\item[$\bullet$]
We propose a novel approach to revise the standard noise prediction network by utilizing the prompt interpolation, which progressively updates the text embedding toward given target prompts during the reverse process of diffusion models.
%\vspace{1mm}
%propose a novel approach to replace the denoising network output in the reverse process formulation with a new output }
%We propose a novel approach to replace the denoising network output in the reverse process formulation with a new estimated output for text-to-driven image-to-image translation}
%We propose a novel method to revise the original denoising process for text-driven image-to-image translation; the proposed method can be conveniently incorporated into existing methods with negligible computational overhead.} 
%to enhance the original denoising process designed for text-driven image-to-image translation.}

\item[$\bullet$] We formulate the proposed noise prediction network using two terms, which is coherent to the conceptual procedure of our task. One is the standard noise prediction term given the source image and the source prompt to reconstruct the overall structure and background, and the other is a new correction term using the progressive prompt interpolation to selectively modify regions of interest. 
%\vspace{1mm} %We derive the new output to adopt the }
%For our denoising process, we employ the denoising network output computed from the source latent and source prompt as inputs for making the target image close to the source image while an additional noise correction term is formulated based on the proposed prompt interpolation for aligning the target image with the target prompt.}
%For the new denoising process, we employ the denoising network output computed from the source latent and source prompt as inputs for making the target image close to the source image while an additional noise correction term is utilized for aligning the target image with the target prompt.}

\item[$\bullet$] Experimental results demonstrate that our proposed method achieves remarkable translation results with time-efficient inference and improves the performance consistently when combined with existing methods.
%We empirically demonstrate that the proposed algorithm outperforms the existing methods .
\end{itemize}

% REMOVED DUE TO limited pages. 
The rest of our paper is organized as follows. 
Section~\ref{sec:related} reviews the related work about text-driven image-to-image translation based on diffusion models. 
Section~\ref{sec:prelim} describes the standard DDIM-based text-driven image-to-image translation algorithm, and Section~\ref{sec:framework} presents our approach. 
Our experimental results are provided in Section~\ref{sec:experiments}, and we finally conclude our paper in Section~\ref{sec:conclusion}.

 % FINISH
% !TEX root = ./../main.tex
\section{Related Work}
\label{sec:related}
This section discusses previous works about diffusion-based text-to-image generation and text-driven image-to-image editing approaches.

\subsection{Text-to-Image Generation based on Diffusion Models}
\label{subsec:t2idiffusion_model}
Diffusion-based text-to-image generation models~\cite{rombach2022high, aditya2022dalle2, saharia2022photorealistic} are typically trained on large-scale training datasets with image-caption pairs.
Motivated by two-stage frameworks~\cite{van2017neural, esser2021taming}, Stable Diffusion~\cite{rombach2022high} projects input images onto a low-dimensional space using a pretrained autoencoder and a diffusion model learns to generate the low-dimensional features conditioned text embeddings given by a text encoder. 
DALLE-2~\cite{aditya2022dalle2} first learns a prior model to estimate CLIP~\cite{radford2021learning} image embeddings based on text captions and then employs a decoder to synthesize output images given the image features and their corresponding text captions.
In contrast, Imagen~\cite{saharia2022photorealistic} utilizes language models~\cite{raffel2020exploring} to extract text features and learns text-to-image diffusion models to generate images conditioned on the text embeddings.

%To compare the proposed method with existing image-to-image translation methods based on the text-to-image diffusion models, 

\iffalse
In a nutshell, in the forward process, diffusion probabilistic models recursively apply Markov Gaussian kernels to the data distribution so that the final distribution becomes a standard multivariate normal distribution.
Although the reverse transition kernel also follows a Markov process, it is intractable to analytically compute the true reverse distribution. 
To sidestep this issue, Denoising Diffusion Probabilistic Models (DDPM)~\cite{ho2020denoising}
%Gaussian distribution with zero mean  
\fi

\subsection{Text-Driven Image Editing based on Diffusion Models}
\label{subsec:i2i}
The goal of text-driven image-to-image translation is to edit the specific regions in a source image to align a resulting target image with the target prompt while preserving the remaining parts.
Existing text-driven image editing methods~\cite{kim2022diffusionclip, brooks2023instructpix2pix, kawar2023imagic, hertz2023prompt, tumanyan2023plug, parmar2023zero} based on diffusion models are typically divided into two groups depending on whether they require an additional training or not.
For example, DiffusionCLIP~\cite{kim2022diffusionclip} fine-tunes a text-to-image diffusion model using the directional CLIP loss~\cite{gal2022stylegan} for fidelity and the identity loss for preserving the background.
Imagic~\cite{kawar2023imagic} optimizes a pretrained diffusion model to reconstruct the source images conditioned on its predicted source text embedding while generating target images based on the interpolation between predicted source text embeddings and target text embeddings.
%based on the interpolation between source and target CLIP text embeddings.

On the other hand, training-free image-to-image translation approaches~\cite{hertz2023prompt, tumanyan2023plug, parmar2023zero} revise the reverse process of pretrained diffusion models. 
For instance, Prompt-to-Prompt~\cite{hertz2023prompt} and Plug-and-Play~\cite{tumanyan2023plug} inject the internal representations of source image---in the forms of cross-attention maps~\cite{hertz2023prompt} or self-attention maps (and simple feature maps)~\cite{tumanyan2023plug}---into the target generation module.
Pix2Pix-Zero~\cite{parmar2023zero} optimizes target latents by aligning the internal representations corresponding to the target and source latents and concurrently generates images with the optimized target latents using the original reverse process. 
Besides, diffusion-based image reconstruction techniques such as Null-text Inversion~\cite{mokady2023null} and Negative-prompt Inversion~\cite{miyake2023negative} can be combined with existing image-to-image translation methods to improve performance, but they are not standalone translation methods.

The proposed approach revises the reverse process of diffusion models without any modification of the text-to-image diffusion backbones. 
Different from existing frameworks~\cite{hertz2023prompt, tumanyan2023plug, parmar2023zero}, we propose a simple but effective method to adjust the noise prediction network for text-driven image-to-image translation.
Since our algorithm is orthogonal to existing methods, we empirically investigate the potential of our approach for performance improvement by combining it with the existing methods.

%Since our algorithm is orthogonal to previous methods, their performance is further enhanced when combined with the proposed method, as validated in Section~\ref{sec:experiments}.
 % CHECK AGAIN!
% !TEX root = ./../main.tex
\section{Text-Driven Image-to-Image Translation}
\label{sec:prelim}
This section describes the standard DDIM-based text-driven image editing approach, which consists of two deterministic processes: the inversion of a source image and the translation to the target domain.

\subsection{Inference of Latent Variables for Source Images}
%Inversion Process of Source Images
\label{subsec:inversion}
Denoising diffusion probabilistic Models (DDPM)~\cite{sohl2015deep, ho2020denoising} assume a Markovian stochastic process with Gaussian transition kernels, where $\bx_0$ is a random variable for an image and $(\bx_1, \bx_2, \cdots ,\bx_T)$ denotes a sequence of latent variables representing intermediate outputs in a diffusion process.
Instead of using DDPM, existing text-driven image-to-image translation methods often rely on the deterministic DDIM inference~\cite{song2021denoising} to reduce the number of inference steps without sacrificing the quality of generated images. 
Utilizing the denoising network denoted by $\epth(\cdot, \cdot, \cdot)$ which is parametrized with the U-Net architecture~\cite{ronneberger2015u}, the forward process of DDIM is formally given by
%a sequence of latent variables, $\bx_1, \bx_2, \cdots \bx_T$ the stochastic process satisfies the Markov process with Gaussian transition kernels with the data \bx_0$ the random variable of the data.
% follows a Gaussian distribution with Ma   
%
\begin{align}
	\label{eq:DDIM_forward}
	\bxsrc_{t+1}= \sqrt{\atpone} \fth(\bxsrc_{t}, t, \bysrc) + \sqrt{1-\atpone} \epth(\bxsrc_t, t, \bysrc),  
\end{align}
where $\bxsrc_t$ is a source latent at a time step $t$, $\bysrc$ is the CLIP text embedding of the source prompt $\psrc$, and $\at$ is a constant decreasing monotonically over time. 
From the above equation, $\fth(\cdot, \cdot, \cdot)$ is derived as 
\begin{align}
	\fth(\bx_{t}, t, \by) = \frac{\bx_t - \sqrt{1-\at} \epth(\bx_t, t,\by )}{\sqrt{\at}}.
	\label{eq:f_theta}
\end{align}
Finally, $\bxsrc_T$ is obtained from $\bxsrc_0$ by recursively leveraging the deterministic DDIM forward process as described in Eq.~\eqref{eq:DDIM_forward}, and is adopted for translating to the image in the target domain, $\bxtgt_0$. 
%reconstructing the source image or trans
%the source image, $\bxsrc_0$, is inverted into the latent ,

%

\iffalse
In a nutshell, in the forward process, diffusion probabilistic models recursively apply Markov Gaussian kernels to the data distribution so that the final distribution becomes a standard multivariate normal distribution.
Although the reverse transition kernel also follows a Markov process, it is intractable to analytically compute the true reverse distribution. 
To sidestep this issue, Denoising Diffusion Probabilistic Models (DDPM)~\cite{ho2020denoising}
\fi

\subsection{Reverse Process of Target Images}
\label{subsec:generative}
By simply setting $\bxtgt_T$ equal to $\bxsrc_T$, one can synthesize the target image using the following DDIM process~\cite{song2021denoising}: 
\begin{align}
\label{eq:naive_target}
\bxtgt_{t-1} =  \sqrt{\atmone} \fth(\bxtgt_{t}, t, \bytgt) + \sqrt{1-\atmone} \epth(\bxtgt_t, t, \bytgt),
\end{align}
where $\bxtgt_t$ is a target latent and $\bytgt$ is a CLIP feature of a target prompt $\ptgt$.
However, the starting point of the reverse process, $\bxtgt_T ( = \bxsrc_T)$, is different from its true position $\mathbf{x}_T^\text{tgt*}$.
Therefore, the na\"ive reverse process often fails to generate desired images in the target domain.
The goal of our approach is to reroute the reverse process to compensate for its wrong initialization and successfully generate target images without additional training.

%\bhr{Although the generated target image is properly aligned with the target prompt, we empirically observe that the deterministic generative process often fails in preserving the structure or background in the source image.}

\begin{figure*}[t!]
	\centering
	%\vspace{-5pt}
	%\vspace{0.2cm}
	\includegraphics[width=1.0\linewidth]{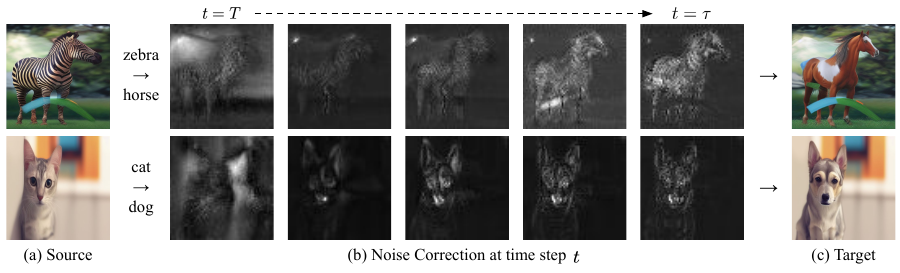}
	%\vspace{-5pt}
	\caption{Visualization of the progressively updated noise correction term $\Delta \epth(\bxtgt, t, \by_t)$ over time for each pair of source and target images.}
	%Qualitative comparisons between the proposed method and state-of-the-art methods~\cite{hertz2023prompt, tumanyan2023plug, parmar2023zero} on real images sampled from the LAION-5B dataset~\cite{schuhmann2022laion} using the pretrained Stable Diffusion~\cite{rombach2022high}.}
%Qualitative results of Pix2Pix on the Cityscapes dataset.
% ``Output" denotes the ground-truth and ``Original" represents the uncompressed generator.}
%\vspace{0.3cm}
\label{fig:visualization_noise_correction}
\end{figure*}
%

%CVPR_MaskVisualization.pdf
% !TEX root = ./../main.tex

\section{Our Approach}
\label{sec:framework}
This section discusses how to improve the quality of translated images for text-driven image-to-image translation.

\subsection{Overview}
% Problem of DDIM Translation
\label{subsec:overview}
%\bhr{As we mentioned in Section~\ref{subsec:generative}, the original DDIM reverse process with the noise prediction $\epth(\bxtgt_t, t, \bytgt)$ results in a generated image $\bxtgt_0$ with significant changes in the background or structure of the source image.}
One of the reasons for poor image-to-image translation quality in na\"ive approaches is the abrupt transition of text embedding from $\bysrc$ to $\bytgt$ at the early stage in the reverse process.
To address this issue, we formulate a noise prediction strategy for the text-driven image-to-image translation by progressively updating the text prompt embedding via time-dependent interpolations of the source and target prompt embeddings.
We derive the revised version of the reverse process and introduce a correction term to update the convergence trajectory conditioned on the target prompt.
Algorithm~\ref{alg:pm} presents the detailed procedure of the proposed method. 
We refer to the proposed algorithm as Prompt Interpolation-based Correction (PIC).

%for the first few steps while we do not modify the original reverse process for the remaining steps.
%Since diffusion models synthesize high-level concepts in the early steps while the fine imperceptible details are refined during the remaining steps as mentioned in~\cite{choi2022perception}.
%Therefore, for the remaining steps, we just perform the reverse DDIM process in Eq.~\eqref{eq:naive_target}. 
%make our new revised noise prediction $\hepth(\bxtgt_t, t, \bytgt)$

\subsection{Noise Correction}
%Modified Noise Prediction}
\label{subsec:mnoise}
To preserve the original structure or background in a source image, we compute a mixture representation of $\bysrc$ and $\bytgt$, which is given by
\begin{align}
%\by_t = \beta_t \cdot \bysrc + (1 - \beta_t) \cdot \bytgt,
\by_t = h( \bysrc, \bytgt, t ),
\label{eq:prompt_interpolation}
\end{align}
where $h(\cdot, \cdot, \cdot)$ is an interpolation function with a time-dependent mixing coefficient $\beta_t$, which will be discussed in Section~\ref{subsec:text_embedding_interpolation}.
Then, we replace the original noise prediction network $\epth(\bxtgt_t, t, \bytgt)$ with a new one, $\hepth(\bxtgt_t, t, \bytgt)$, which is given by
\begin{align}
\hepth(\bxtgt_t, t, \bytgt) := \epth(\bxsrc_t, t, \bysrc) + \gamma \Delta \epth(\bxtgt_t, t, \by_t),
\label{eq:new_noise_prediction}
\end{align}
where $\gamma$ is a hyperparameter and $\Delta \epth(\bxtgt_t, t, \by_t)$ is a correction term with the interpolated text prompt embedding
$\by_t$. 
%that ensures the target image to fall within the target domain.
%making the target image to lie in the target domain.
%estimated based on $\epth(\bxsrc_t, t, \bysrc)$.
%Based on the new noise prediction, 
%
In this formulation, $\epth(\bxsrc_t, t, \bysrc)$ enables our approach to preserve the structure or background of a source image $\bxsrc_0$ while the additional correction term facilitates the alignment of the generated image to the target domain.
%The detailed formulation of the correction term will be discussed in the next subsection.
%source-noise correction and text interpolation 

%can address this problem.
%\\

Conceptually, it is desirable for the noise correction term, $\Delta \epth(\bxtgt_t, t, \by_t)$, to only affect the relevant regions to the target prompt while preserving the rest of the source image.
The formal definition of the correction term $\Delta \epth(\bxtgt_t, t, \by_t)$ is as follows:
\begin{align}
\Delta \epth(\bxtgt_t, t, \by_t):= \epth(\bxtgt_t, t, \by_t) - \epth(\bxtgt_t, t, \bysrc),
\label{eq:correction_term}
\end{align}
where $\by_t$ moves from $\bysrc$ to $\bytgt$ as $t$ decreases.
% where $\by_t$ fades away from $\bysrc$ but moves toward $\bxtgt_t$ as $t$ decreases.
By plugging Eq.~\eqref{eq:correction_term} into Eq.~\eqref{eq:new_noise_prediction}, we obtain the following noise prediction network:
\begin{align}
\label{eq:new_noise_prediction2}
\hepth(\bxtgt_t, t, \bytgt) & = \epth(\bxsrc_t, t, \bysrc) + \gamma \left( \epth(\bxtgt_t, t, \by_t) - \epth(\bxtgt_t, t, \bysrc) \right). 
%\hepth(\bxtgt_t, t, \bytgt) & := \gamma \epth(\bxtgt_t, t, \by_t) + \epth(\bxsrc_t, t, \bysrc) - \gamma \epth(\bxtgt_t, t, \bysrc)
\end{align}

\iffalse
%smooth transition from $\bysrc$ to $\bytgt$ during the reverse process of $\bxtgt$. 
%linear interpolation of the two text embedding between $\bysrc$ and $\bytgt$ so that $\by_t$ approaches $\bysrc$ when $t$ is closer to 0 while $\by_t$ is similar to $\bytgt$ when $t$ is near $T$.
%, which becomes close to $\bysrc$ at time step $T$ while approaches 
%from $\bysrc$ to $\bytgt$. 
%
\begin{align}
\by_t :=  \beta_t \bysrc + (1-\beta_t) \bytgt.
\label{eq:text_interp}
\end{align}
%
%In Eq.~\eqref{eq:text_interp}, we simply set $\beta_t$ to $\frac{T-t}{T}$. 
\fi

Our intuition behind the noise correction term is that the noise prediction given the target latent and the progressively interpolated text embedding effectively makes up for the gap between the unknown true initialization, $\mathbf{x}_T^\text{tgt*}$ and its trivial surrogate, $\bxtgt_T ( = \bxsrc_T)$. 
We observe that this correction term is particularly helpful at the early stage of the reverse process and is not necessarily required for the rest of the iterations.
%is different only at the region of interest compared with that computed from the target latent and the source text embedding. 
%Based on the noise correction term, we expect it facilitates the image-to-image translation. 
%Figure~\ref{} supports the intuition since 
\cref{fig:visualization_noise_correction} supports our intuition by visualizing the noise correction term during the reverse process; it gradually highlights the regions to be updated while the background area
is set to negligible values.
%We empirically confirm that the noise correction term preserves the background region since it is constantly set to 0 during the generative process as visualized in Figure~\ref{}.}
%Also, we empirically observe that the translation based on our revised prediction successfully edits the object region to meaningfully align with the target domain while preserving the background.

%
\begin{algorithm}[t]
  \caption{Target image generation by PIC}
  \label{alg:pm}
  %\centering
  \begin{algorithmic}[1]
  \State{\bf Input:} source image $\bxsrc_0$, source prompt embedding $\bysrc$,  target prompt embedding $\bytgt$, hyperparameters $\beta$, $\gamma$, $\tau$  
   \For{$t \gets 0, \cdots , T-1$}
		 \State Compute $\epth(\bxsrc_t, t, \bysrc)$ and obtain $\bxsrc_{t+1}$ by Eq.~\eqref{eq:DDIM_forward} while saving $\epth(\bxsrc_t, t, \bysrc)$%and Store $\bxsrc_{t+1}$ %and $\Asrc_t$
		\EndFor
  \State $\bxtgt_T \gets \bxsrc_T$
  \For{$t \gets T, \cdots , T-\tau+1$}
  		 \State Obtain $\by_t$ based on $\bysrc$ and $\bytgt$ using Eq.~\eqref{eq:mixup_word_replacement} or Eq.~\eqref{eq:mixup_adding_phrasess} depending on the given task 
		 \State Compute $\epth(\bxtgt_t, t, \by_t)$ and $\epth(\bxtgt_t, t, \bysrc)$ 
		 %\State Obtain the noise correction term $\Delta \epth(\bxtgt, t, \by_t)$ using Eq.~\eqref{eq:correction_term}
		 \State Obtain the revised model $\hepth(\bxtgt_t, t, \bytgt)$ using Eq.~\eqref{eq:new_noise_prediction2}
		 \State Obtain $\bxtgt_{t-1}$ using Eq.~\eqref{eq:naive_target} by replacing $\epth(\bxtgt_t, t, \bytgt)$ with $\hepth(\bxtgt_t, t, \bytgt)$  %and Store $\bxsrc_{t+1}$ %and $\Asrc_t$
		\EndFor
	  \For{$t \gets T-\tau, \cdots , 1$}
	  	\State Obtain $\bxtgt_{t-1}$ using Eq.~\eqref{eq:naive_target}
	  \EndFor
		\State {\bf Output:} target image $\bxtgt_0$
  \end{algorithmic}
\end{algorithm}

\subsection{Prompt Interpolation}
%Text Embedding Interpolation}
\label{subsec:text_embedding_interpolation}
%Since our text interpolation varies depending on the relationship between the source and target prompts, we describe it for the two tasks: word swap and adding word phrases.
We now describe the proposed prompt interpolation strategies with the source and target embeddings, designed for the slightly different two tasks of interest: word replacement and adding phrases.

\subsubsection{Word replacement}
For word replacement, we consider the scenario that the tokens in the source prompt are replaced by other ones.
For example, in the case of `zebra $\rightarrow$ horse', if the source prompt is \textit{`A zebra is lying on the grass.'}, the target prompt becomes \textit{`A horse is lying on the grass' } by replacing {`zebra'} with {`horse'}.
%becomes \textit{`A horse is lying on the grass' } with a replacement of \textit{`zebra'} with \textit{`horse'}. 
In this task, our simple linear prompt interpolation is given by
%\ms{$\by_t[\ell] \in \mathbb{R}^{D}$} f}
%, the mixed CLIP embedding corresponding to the $\ell^\text{th}$ token, which is given by  }
%
\begin{align}
\by_t[\ell] = \beta_t \bytgt[\ell] + (1-\beta_t) \bysrc[\ell], 
\label{eq:mixup_word_replacement}
\end{align}
where $\ell$ is a token index and the time-dependent coefficient $\beta_t$ is set to
\begin{align}
\beta_t:= \beta + (1-\beta) \times \frac{T-t}{T},
\label{eq:def_bt}
\end{align}
%
%s set to $\beta + (1-\beta) \frac{T-t}{T}$ with a hyperparameter $\beta$.
%Note that the interpolation has no effect on the embeddings corresponding to the tokens located before the exchanged one since the two source and target embeddings matched with the tokens are the same. 
where $\beta$ is an initialization value between 0 and 1.
Note that the interpolated embedding is progressively updated starting from the source prompt embedding to the target prompt embedding during the reverse process.

\subsubsection{Adding phrases}
We consider another task that involves the addition of tokens. 
For instance, in the case of `dog $\rightarrow$ dog with glasses', if the source prompt is \textit{`A dog is lying on the grass'}, then the target prompt becomes \textit{`A dog with glasses is lying on the grass'}.
In this task, we have to match tokens between the source and target prompt embeddings for prompt interpolation, which is given by
\begin{align}
\label{eq:mixup_adding_phrasess}
 \by_t[\ell] =  \begin{cases}
\bysrc[\ell], &\text{if } \ell < \ell_s \\
\bytgt[\ell], &\text{if } \ell_s \leq \ell \leq \ell_f  \\
\beta_t \bytgt[\ell] + (1-\beta_t) \bysrc[\ell - \ell_f + \ell_s], &\text{if } \ell > \ell_f  
 %\ell_s \leq \ell \leq \ell_f \\ %[\ell_s, \ell_f] \\
\end{cases}
\end{align}
where $n\,(= \ell_f - \ell_s + 1)$ tokens are inserted at the $\ell_s^\text{th}$ position of the source prompt and $\beta_t$ is defined in Eq.~\eqref{eq:def_bt}. 
%the phrase with $(\ell_f - \ell_s)$ words are inserted at the $\ell_s^\text{th}$ position of the source prompt. 
Note that this strategy interpolates the embeddings of the source and target prompts from the next token positions of the added phrase\footnote{Our prompt interpolation strategy for adding phrases can be generalized to the cases where phrases are removed.}
%removing phrases.}. 
%The intention for utilizing adaptive interpolation is to merge the token embedding of the target prompt with the embedding derived from a meaningfully matched token in the source prompt.
%In the case of the target tokens absent in the source prompt, we simply employ the corresponding target prompt embedding without considering the interpolation.

%\subsection{Extensions}

\subsection{Integration into Existing Methods}
\label{subsubsec:integration}
The proposed technique can be conveniently incorporated into state-of-the-art methods for diffusion-based image-to-image translation such as Prompt-to-Prompt ~\cite{hertz2023prompt}, Plug-and-Play~\cite{tumanyan2023plug}, and Pix2Pix-Zero~\cite{parmar2023zero}.
%When we do not utilize the proposed modified prediction, we employ the previous algorithms for the remaining time steps.
The algorithm-specific noise prediction network, $\hepth(\bxtgt_t, t, \bytgt)$, derived from Eq.~\eqref{eq:new_noise_prediction} is expressed as 
\begin{align}
\label{eq:algo_new_noise_prediction}
\hepth^{\text{alg}} & (\bxtgt_t, t, \bytgt):= \epth(\bxsrc_t, t, \bysrc) + \gamma \Delta \epth^\text{alg}(\bxtgt_t, t, \by_t), 
\end{align}
where $\Delta \epth^\text{alg}(\bxtgt_t, t, \by_t)$ is the noise correction term, specific to the individual translation algorithms~\cite{hertz2023prompt, tumanyan2023plug, parmar2023zero}. %discussed in the following subsections. 
%On the other hand, during the remaining steps, since we do not use the proposed modified prediction as we mentioned, we simply adopt the previous methods without any modification.%
The rest of this subsection discusses how to obtain the new noise correction term $\Delta \epth^\text{alg}(\bxtgt_t, t, \by_t)$ for each algorithm.

\subsubsection{Prompt-to-Prompt~\cite{hertz2023prompt}}
The extension of the proposed prompt interpolation technique to Prompt-to-Prompt is simple.
%During the computation of $\epth(\bxtgt_t, t, \by_t)$ for the noise correction term in Eq.~\eqref{eq:correction_term}, 
During the reverse process, Prompt-to-Prompt replaces the cross-attention and self-attention maps in $\epth(\bxtgt_t, t, \bytgt)$ with the matching attention maps in $\epth(\bxsrc_t, t, \bysrc)$.
Different from the vanilla Prompt-to-Prompt, our extension 
replaces the attention maps in $\epth(\bxtgt_t, t, \by_t)$ to ones in $\epth(\bxsrc_t, t, \bysrc)$, instead of $\epth(\bxtgt_t, t, \bytgt)$.

\iffalse
Then, we define $\Delta \epth^{\text{PtP}} (\bxtgt_t, t, \by_t)$ as follows:
%the noise correction term $\Delta \epth^{\text{PtP}} (\bxtgt, t, \by_t)$ for the integration of the proposed method and Prompt-to-Prompt~\cite{hertz2023prompt} is defined as follows:  
%
\begin{align}
\Delta \epth^{\text{PtP}}(\bxtgt_t, t, \by_t):= \epth^\text{PtP}(\bxtgt_t, t, \by_t) - \epth(\bxtgt_t, t, \bysrc),
\label{eq:PtP_correction_term}
\end{align}
%
where $\epth(\bxtgt_t, t, \by_t)$ in the original term is replaced with $\epth^\text{PtP}(\bxtgt_t, t, \by_t)$. 
%given by the cross-attention injection based on the proposed text embedding interpolation. 
\fi

\subsubsection{Plug-and-Play~\cite{tumanyan2023plug}}
Plug-and-Play performs the reverse process with the substitution of the self-attention maps and the intermediate feature maps in the denoising network $\epth(\bxsrc_t, t, \bysrc)$ for those obtained from $\epth(\bxtgt_t, t, \bytgt)$.
As in our extension to Prompt-to-Prompt, we use $\epth(\bxtgt_t, t, \by_t)$ instead of $\epth(\bxtgt_t, t, \bytgt)$ to compute the attention and feature maps for the replacements.
%The rest of the procedure is identical to the original Plug-and-Play implementation.

\iffalse
%Unlike Plug-and-Play, our method utilizes the injection during the computation of $\epth(\bxtgt_t, t, \by_t)$ rather than $\epth(\bxtgt_t, t, \bytgt)$.
After the injection process, we have the enhanced noise prediction $\epth^\text{PnP}(\bxtgt_t, t, \by_t)$, which alters the original noise correction term as 
%
\begin{align}
\Delta \epth^{\text{PnP}}(\bxtgt_t, t, \by_t):= \epth^\text{PnP}(\bxtgt_t, t, \by_t) - \epth(\bxtgt_t, t, \bysrc).
\label{eq:PnPcorrection_term}
\end{align}
%
%where $\Delta \epth^{\text{PnP}}(\bxtgt, t, \by_t)$ is the modified noise correction term given by the proposed method combined with Plug-and-Play~\cite{tumanyan2023plug}.
\fi

\subsubsection{Pix2Pix-Zero~\cite{parmar2023zero}}
For the reverse process, Pix2Pix-Zero obtains the target latent $\hbxtgt_t$ by taking a gradient step from $\bxtgt_t$ using the cross-attention guidance loss, which aims to align the cross attention maps in the denoising network given the source and the target latents.
%two different noise predictions.
The optimized target latent is given by
\begin{align}
\hbxtgt_t = \bxtgt_t - \lambda_{\text{xa}} \nabla_{\bxtgt_t} \| \Mtgt_t - \Msrc_t \|^2_F. 
\label{eq:addon_p2p}
\end{align} 
where $\Mtgt_t$ and $\Msrc_t$ denote the cross attention maps in $\epth(\bxtgt_t, t, \by_t)$ and $\epth(\bxsrc_t, t, \bysrc)$. Respectively, $\lambda_{\text{xa}}$ is a hyperparameter, and $\| \cdot \|_F$ indicates the Frobenius norm. Note that vanilla Pix2Pix-Zero obtains $\Mtgt_t$ from $\epth(\bxtgt_t, t, \bytgt)$.
Therefore, the noise correction term specific to Pix2Pix-Zero, is given by
\begin{align}
\Delta \epth^{\text{P2P}}(\bxtgt_t, t, \by_t):= \epth(\hbxtgt_t, t, \by_t) - \epth(\hbxtgt_t, t, \bysrc),
\label{eq:PXPcorrection_term}
\end{align}
where $\bxtgt_t $ is replaced by $\hbxtgt_t$ from Eq.~\eqref{eq:correction_term}.
%standard equation.

% !TEX root = ./../main.tex
\section{Experiments}
\label{sec:experiments}
We evaluate the performance of our approach, PIC, in comparison with the state-of-the-art training-free diffusion-based image-to-image translation methods~\cite{hertz2023prompt, tumanyan2023plug, parmar2023zero}. 
We identify the $250$ most relevant images for the desired source domain given a task, based on their CLIP similarities, and use them as inputs for image-to-image translation methods to be tested in the task.
Note that the algorithm integrating PIC is denoted by by `[Algorithm Name] + PIC'.

\begin{table}[t!]
	\centering
	\caption{
		Quantitative comparisons of PIC with Prompt-to-Prompt~\cite{hertz2023prompt}, Plug-and-Play~\cite{tumanyan2023plug}, and Pix2Pix-Zero~\cite{parmar2023zero} on images sampled from the LAION-5B dataset~\cite{schuhmann2022laion} using the pretrained Stable Diffusion~\cite{rombach2022high} backbone. 
		Black and red bold-faced numbers denote the best and second-best performances within each row for each metric. 
	}
	\setlength\tabcolsep{2pt} 
	%\hspace{-3mm}
	\scalebox{0.75}{
		%\resizebox{\linewidth}{!}{
			\begin{tabular}{l | ccc | ccc | ccc | ccc}
				%\begin{tabular}{l ccc ccc ccc ccc ccc}
				%\toprule 
				%\hline
				\multirow{2}{*}{Task} 
				%& \multicolumn{3}{c}{DDIM}
				& \multicolumn{3}{c|}{PtP}
				& \multicolumn{3}{c|}{PnP}
				& \multicolumn{3}{c|}{P2P} 
				& \multicolumn{3}{c}{PIC (Ours)} \\ 
				%& \multicolumn{3}{c}{Avg} \\ %\\
				
				%\cline{2-4} \cline{5-7} \cline{8-10} \cline{11-13}
				%\cline{2-4} \cline{5-7} \cline{8-10} \cline{11-13} \cline{14-16}
				& CS ($\uparrow$)
				& BD ($\downarrow$)
				& SD ($\downarrow$)

				& CS ($\uparrow$)
				& BD ($\downarrow$)
				& SD ($\downarrow$)
				
				& CS ($\uparrow$)
				& BD ($\downarrow$)
				& SD ($\downarrow$)

				& CS ($\uparrow$)
				& BD ($\downarrow$)
				& SD ($\downarrow$)
				
				%& CS ($\uparrow$)
				%& BD ($\downarrow$)
				%& SD ($\downarrow$)
				
				\\ 
				\cline{1-13}
				%\cline{1-16}
				
				{dog $\rightarrow$ cat}
				& \bred{0.290} & \bred{0.076} & 0.038 
				& \black{0.293} & 0.100 & \bred{0.032} 
				& 0.281 & 0.127 & 0.099
				& \black{0.293} & \black{0.045} & \black{0.031} \\
				%& \black{0.293} & \black{0.045} & \black{0.031}\\

				{cat $\rightarrow$ dog}
				% & ? & ? & ?
				& \bred{0.288} & \bred{0.095} & \bred{0.042}
				& \black{0.291} & 0.099 & \black{0.033}
				& 0.282 & 0.100 & 0.054
				& \bred{0.288} & \black{0.057} & \black{0.033} \\
				%& \black{0.293} & \black{0.045} & \black{0.031}\\

				{horse $\rightarrow$ zebra}
				% & ? & ? & ?
				& 0.320	& \bred{0.133}	& \bred{0.042}
				& \black{0.333}	& 0.158	& \bred{0.042}
				& 0.323	& 0.193	& 0.078
				& \bred{0.324}	& \black{0.085}	& \black{0.037} \\
				%& \black{0.293} & \black{0.045} & \black{0.031}\\

				{zebra $\rightarrow$ horse}
				% & ? & ? & ?
				& 0.291	& 0.183	& 0.051
				& \black{0.299}	& \bred{0.152}	& \black{0.043}
				& 0.282	& 0.216	& 0.104
				& \bred{0.292}	& \black{0.126}	& \bred{0.050} \\
				%& \black{0.293} & \black{0.045} & \black{0.031}\\

				{tree $\rightarrow$ palm tree}
				& \black{0.315}	& 0.147	& 0.045
				& \bred{0.314}  & \bred{0.122}	& \bred{0.039}
				& \bred{0.314}  & 0.129	& 0.046 
				& \bred{0.314}	& \black{0.085}	& \black{0.036} \\
				% & \black{0.293} & \black{0.045} & \black{0.031}\\
				
				{dog $\rightarrow$ dog w/glasses}
				& 0.310	& \bred{0.041}	& 0.020
				& 0.302	& 0.087	& 0.025 
				& \black{0.322}	& 0.050	& \black{0.015} 
				& \bred{0.312}	& \black{0.026}	& \bred{0.016} \\
				%& \black{0.293} & \black{0.045} & \black{0.031} \\

				\hline
				{Average}
				& 0.302	& \bred{0.113}	& 0.040
				& \black{0.305}	& 0.120	& \bred{0.036} 
				& 0.301	& 0.136	& 0.066 
				& \bred{0.304}	& \black{0.071}	& \black{0.034} \\
				%& \black{0.293} & \black{0.045} & \black{0.031} \\
				
				\hline
				%\bottomrule 
			\end{tabular}
			%}
	}
	\label{tab:cmp_baselines} 
\end{table} %TODO: Reshape this table
\begin{table}[t!]
	\centering
	\caption{
		Quantitative comparisons of the proposed method with Prompt-to-Prompt~\cite{hertz2023prompt} on images sampled from the LAION-5B dataset~\cite{schuhmann2022laion} using the pretrained Stable Diffusion~\cite{rombach2022high}.
		Out technique is integrated into Prompt-to-Prompt and the results of Prompt-to-Prompt are obtained from \cref{tab:cmp_baselines}.
		Black bold-faced numbers represent better performance on each metric between two approaches. 
	}
	\setlength\tabcolsep{8pt} 
	\scalebox{0.75}{
		%\resizebox{\linewidth}{!}{
			\begin{tabular}{l | ccc | ccc }
				%\toprule 
				%\hline
				\multirow{2}{*}{Task} 
				& \multicolumn{3}{c|}{PtP}
				& \multicolumn{3}{c}{PtP + PIC (Ours)}\\
				
				%\cline{2-4} \cline{5-7} %\cline{8-10} \cline{11-13}
				
				& CS ($\uparrow$)
				& BD ($\downarrow$)
				& SD ($\downarrow$)

				& CS ($\uparrow$)
				& BD ($\downarrow$)
				& SD ($\downarrow$)
				
				\\ 
				\cline{1-7}
				
				{dog $\rightarrow$ cat}
				& \black{0.290} & 0.076 & 0.038 
				& 0.283 & \black{0.051} & \black{0.021}  \\
				%& 0.281 & 0.127 & 0.099
				%& \black{0.293} & \black{0.046} & \black{0.031} \\

				{cat $\rightarrow$ dog}
				& 0.288 & 0.095 & 0.042 
				& \black{0.291} & \black{0.052} & \black{0.027} \\
				%& 0.282 & 0.100 & 0.054
				%& \red{0.288} & \black{0.060} & \red{0.034} \\

				{horse $\rightarrow$ zebra}
				& \black{0.320}	& 0.133	& 0.042
				& 0.292	& \black{0.071}	& \black{0.018} \\
				%& 0.323	& 0.193	& 0.078
				%& \red{0.324}	& \black{0.133}	& \black{0.036} \\

				{zebra $\rightarrow$ horse}
				& \black{0.291}	& 0.183	& 0.051
				& 0.290	& \black{0.131}	& \black{0.034} \\
				%& 0.282	& 0.216	& 0.104
				%& \red{0.293}	& \black{0.086}	& \red{0.051} \\

				{tree $\rightarrow$ palm tree}
				& \black{0.315}	& 0.147	& 0.045
				& 0.301	& \black{0.070}	& \black{0.026} \\
				%& 0.314	& -	& 0.046 
				%& 0.314	& -	& 0.043 \\
				
				{dog $\rightarrow$ dog w/glasses}
				& \black{0.310}	& 0.041	& 0.020
				& 0.301	& \black{0.038}	& \black{0.011} \\
				%& \black{0.322}	& 0.050	& \black{0.015} 
				%& \red{0.316}	& \black{0.037}	& 0.022 \\
				
				\hline
				{Average}
				& \black{0.302}	& 0.113	& 0.040
				& 0.295	& \black{0.069}	& \black{0.023} \\
				
				\hline
				%\bottomrule 
			\end{tabular}
			%}
	}
	\label{tab:addon_ptp}
\end{table}

\subsection{Implementation Details}
We implement the proposed method using the publicly available code of Pix2Pix-Zero (P2P)\footnote{\url{https://github.com/pix2pixzero/pix2pix-zero}}.
We integrate PIC into the existing techniques---Prompt-to-Prompt (PtP)\footnote{\url{https://github.com/google/prompt-to-prompt}}, Plug-and-Play (PnP)\footnote{\url{https://github.com/MichalGeyer/plug-and-play}} and Pix2Pix-Zero (P2P)---using their official codes.
% Also, we integrate our algorithm into the existing techniques---Prompt-to-Prompt (PtP)\footnote{\url{https://github.com/google/prompt-to-prompt}} and Plug-and-Play (PnP)\footnote{\url{https://github.com/MichalGeyer/plug-and-play}}---using their official codes.
To accelerate the text-driven image-to-image translation process, the inference time steps for the forward and reverse processes are set to $50$. For all experiments, \bhr{Stable Diffusion v1.4 is employed as the backbone model.}
During the forward process, we adopt Bootstrapping Language-Image Pretraining (BLIP)~\cite{li2022blip} to generate a source prompt for conditioning the denoising network.   
The target prompt is given by replacing the specific words in the source prompt with the alternatives defined by an assigned task as mentioned in Section~\ref{subsec:text_embedding_interpolation}.
We use the same source and target prompts of all algorithms for the fair comparisons during both the forward and reverse processes. 
Additionally, we adopt classifier-free guidance~\cite{ho2021classifier} following~\cite{hertz2023prompt, tumanyan2023plug, parmar2023zero}. 

In our implementation, $\tau$ and $\gamma$ are set to $25$ and $1.0$, respectively, for all experiments. 
Also, we set $\beta$ to $0.3$ for word replacement tasks (\eg `dog $\rightarrow$ cat' and `horse $\rightarrow$ zebra') while it is set to $0.8$ for adding phrases tasks (\eg `tree $\rightarrow$ palm tree' and `dog $\rightarrow$ dog with glasses').

\begin{table}[t!]
	\centering
	\caption{
		Quantitative comparisons of the proposed method with Plug-and-Play~\cite{tumanyan2023plug} on images sampled from the LAION-5B dataset~\cite{schuhmann2022laion} using the pretrained Stable Diffusion~\cite{rombach2022high}.
		Out technique is integrated into Plug-and-Play and the results of Plug-and-Play are obtained from~\cref{tab:cmp_baselines}.
	}
	%\vspace{-0.1in}
	%\vspace{0.1in}
	\setlength\tabcolsep{8pt} 
	\scalebox{0.75}{
		%\resizebox{\linewidth}{!}{
			\begin{tabular}{l | ccc | ccc}
				%\toprule 
				%\hline
				\multirow{2}{*}{Task} 
				& \multicolumn{3}{c|}{PnP}
				& \multicolumn{3}{c}{PnP + PIC (Ours)}\\
				
				%\cline{2-4} \cline{5-7} %\cline{8-10} \cline{11-13}
				
				& CS ($\uparrow$)
				& BD ($\downarrow$)
				& SD ($\downarrow$)

				& CS ($\uparrow$)
				& BD ($\downarrow$)
				& SD ($\downarrow$)
				
				\\ 
				\cline{1-7}
				
				{dog $\rightarrow$ cat}
				%& \red{0.290} & \red{0.076} & 0.038 
				& \black{0.293} & 0.100 & 0.032  
				& 0.282 & \black{0.092} & \black{0.027} \\
				%& \black{0.293} & \black{0.046} & \black{0.031} \\			

				{cat $\rightarrow$ dog}
				%& \red{0.288} & \red{0.095} & 0.042 
				& \black{0.291} & 0.099 & 0.033 
				& 0.288 & \black{0.083} & \black{0.028} \\
				%& \red{0.288} & \black{0.060} & \red{0.034} \\

				{horse $\rightarrow$ zebra}
				%& 0.320	& \black{0.133}	& \red{0.042}
				& \black{0.333}	& 0.158	& 0.042 
				& 0.317	& \black{0.121}	& \black{0.035} \\
				%& \red{0.324}	& \black{0.133}	& \black{0.036} \\

				{zebra $\rightarrow$ horse}
				%& 0.291	& 0.183	& 0.051
				& \black{0.299}	& 0.152	& 0.043 
				& 0.285	& \black{0.135}	& \black{0.037} \\
				%& \red{0.293}	& \black{0.086}	& \red{0.051} \\

				{tree $\rightarrow$ palm tree}
				%& 0.315	& -	& 0.045
				& \black{0.314}	& 0.122	& 0.039 
				& 0.295	& \black{0.070} & \black{0.024}  \\
				%& 0.314	& -	& 0.043 \\
				
				{dog $\rightarrow$ dog w/glasses}
				%& 0.310	& \red{0.041}	& \red{0.020}
				& \black{0.302}	& 0.087	& 0.025 
				& 0.300	& \black{0.085}	& \black{0.024} \\
				%& \red{0.316}	& \black{0.037}	& 0.022 \\
				
				\hline
				{Average}
				& \black{0.305}	& 0.120	& 0.036 
				& 0.295	& \black{0.098}	& \black{0.029} \\
				
				\hline
				%\bottomrule 
			\end{tabular}

			%}
	}
	\label{tab:addon_pnp}
\end{table}
\begin{table}[t!]
	\centering
	\caption{
		Quantitative comparisons of the proposed method with Pix2Pix-Zero~\cite{parmar2023zero} on images sampled from the LAION-5B dataset~\cite{schuhmann2022laion} using the pretrained Stable Diffusion~\cite{rombach2022high}.
		Out technique is integrated into Pix2Pix-Zero and the results of Pix2Pix-Zero are obtained from~\cref{tab:cmp_baselines}.
	}
	%\vspace{0.1in}
	%\vspace{-0.1in}
	\setlength\tabcolsep{8pt} 
	\scalebox{0.75}{
		%\resizebox{\linewidth}{!}{
			\begin{tabular}{l | ccc | ccc}
				%\toprule 
				%\hline
				\multirow{2}{*}{Task} 
				& \multicolumn{3}{c|}{P2P}
				& \multicolumn{3}{c}{P2P + PIC (Ours)}\\
				
				%\cline{2-4} \cline{5-7} %\cline{8-10} \cline{11-13}
				
				& CS ($\uparrow$)
				& BD ($\downarrow$)
				& SD ($\downarrow$)

				& CS ($\uparrow$)
				& BD ($\downarrow$)
				& SD ($\downarrow$)
				
				\\ 
				\cline{1-7}
				
				{dog $\rightarrow$ cat}
				%& \red{0.290} & \red{0.076} & 0.038 
				%& \black{0.293} & 0.100 & \red{0.032}  \\
				& 0.281 & 0.127 & 0.099
				& \black{0.282} & \black{0.051} & \black{0.017} \\

				{cat $\rightarrow$ dog}
				%& \red{0.288} & \red{0.095} & 0.042 
				%& \black{0.291} & 0.099 & \black{0.033} \\
				& 0.282 & 0.100 & 0.054
				& \black{0.285} & \black{0.056} & \black{0.016} \\

				{horse $\rightarrow$ zebra}
				%& 0.320	& \black{0.133}	& \red{0.042}
				%& \black{0.333}	& 0.158	& \red{0.042} \\
				& \black{0.323}	& 0.193	& 0.078
				& 0.309	& \black{0.070}	& \black{0.016} \\

				{zebra $\rightarrow$ horse}
				%& 0.291	& 0.183	& 0.051
				%& \black{0.299}	& \red{0.152}	& \black{0.043} \\
				& \black{0.282}	& 0.216	& 0.104
				& 0.279	& \black{0.117}	& \black{0.017} \\

				{tree $\rightarrow$ palm tree}
				%& 0.315	& -	& 0.045
				%& ?	& -	& ? \\
				& \black{0.314}	& 0.129	& 0.046 
				& 0.298	& \black{0.047}	& \black{0.014} \\
				
				{dog $\rightarrow$ dog w/glasses}
				%& 0.310	& \red{0.041}	& \red{0.020}
				%& 0.302	& 0.087	& 0.025 \\
				& \black{0.322}	& \black{0.050}	& 0.015 
				& 0.302	& 0.053	& \black{0.011} \\

				\hline
				{Average}
				%& 0.310	& \red{0.041}	& \red{0.020}
				%& 0.302	& 0.087	& 0.025 \\
				& \black{0.301}	& 0.136	& 0.066 
				& 0.293	& \black{0.066}	& \black{0.015} \\

				\hline
				%\bottomrule 
			\end{tabular}
			%}
	}
	\label{tab:addon_p2p}
\end{table}
\begin{figure}[p]
	\centering
	\includegraphics[width=\linewidth]{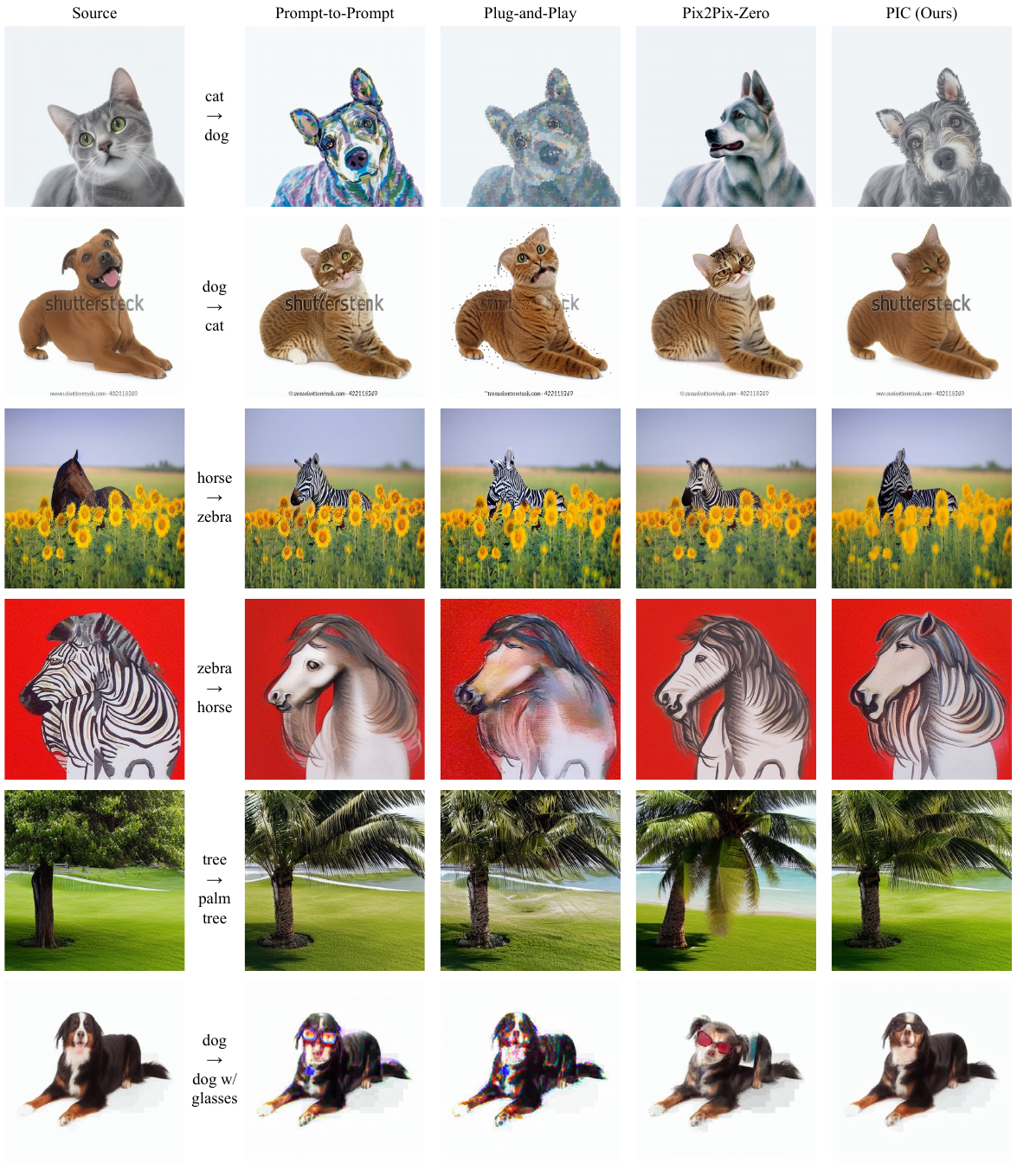}
	%{\scriptsize \sf{A failure case}} 
	\includegraphics[width=\linewidth]{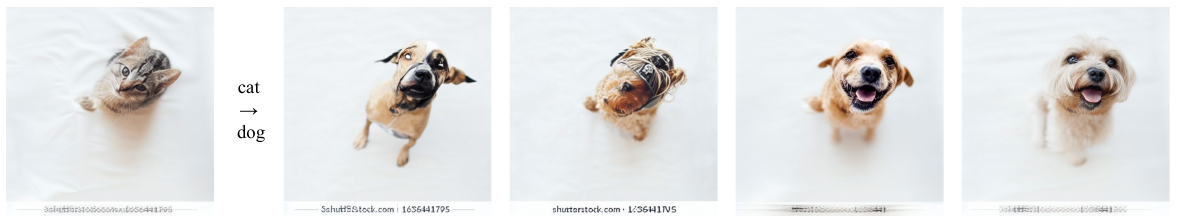}
	%\vspace{-5pt}
	\caption{Qualitative comparisons between PIC and state-of-the-art methods~\cite{hertz2023prompt, tumanyan2023plug, parmar2023zero} on images from LAION-5B~\cite{schuhmann2022laion} using the pretrained Stable Diffusion~\cite{rombach2022high}.
		PIC generates target images with higher-fidelity than others in all tasks. 
		Note that all algorithms fail to preserve pose and texture of the source image in the last task, but PIC still shows a favorable result.
	}
\label{fig:comparisons}
\end{figure}

\begin{figure}
\begin{subfigure}{\linewidth}
	\centering
	\includegraphics[width=1.0\linewidth]{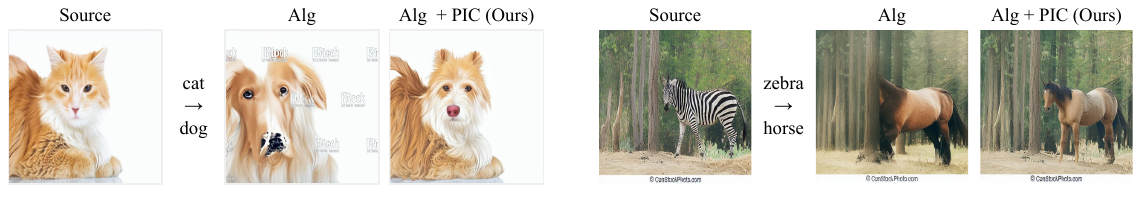}
	%\caption{Prompt-to-Prompt~\cite{hertz2023prompt}}
	%\label{fig:first}
\end{subfigure}
\begin{subfigure}{\linewidth}
	\centering
	\includegraphics[width=1.0\linewidth]{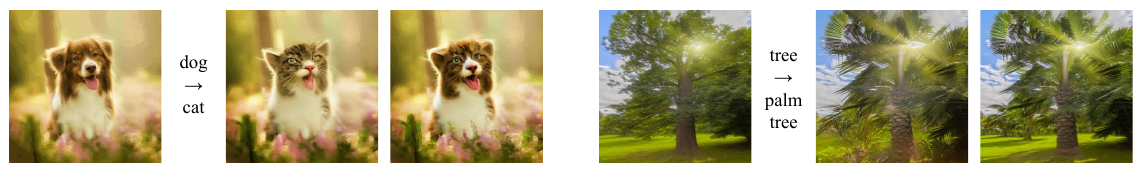}
	%\caption{Plug-and-Play~\cite{tumanyan2023plug}}
	%\label{fig:second}
\end{subfigure}
\begin{subfigure}{\linewidth}
	\centering
	\includegraphics[width=1.0\linewidth]{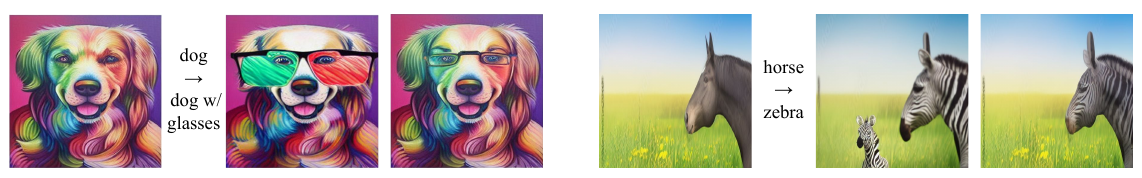}
	%\caption{Pix2Pix-Zero~\cite{parmar2023zero}}
	%\label{fig:third}
\end{subfigure}
%\vspace{3pt}
\caption{Qualitative results of existing state-of-the-art methods and their combinations with PIC based on the pretrained Stable Diffusion~\cite{rombach2022high}: (top) Prompt-to-Prompt~\cite{hertz2023prompt}, (middle) Plug-and-Play~\cite{tumanyan2023plug}, and (bottom) Pix2Pix-Zero~\cite{parmar2023zero}.
The examples are sampled from LAION-5B~\cite{schuhmann2022laion}.
	%The proposed method is effective to preserve the structure or background when combined with the previous frameworks.
}
%Prompt-to-Prompt~\cite{}, Plug-and-Play~\cite{}, Pix2Pix-Zero~\cite{}, and }
\label{fig:addon}
\end{figure}

\subsection{Evaluation Metrics}
For quantitative evaluation, we measure CLIP Similarity~\cite{hessel2021clipscore}, Background Distance, and Structure Distance~\cite{tumanyan2022splicing} following Pix2Pix-Zero~\cite{parmar2023zero}. 
The CLIP similarity (CS) quantifies how well the translated images are aligned with the target prompts using the cosine similarity.
On the other hand, the background distance (BD) calculates the Learned Perceptual Image Patch Similarity (LPIPS) score~\cite{zhang2018unreasonable} between the background regions of the source and translated images.
To identify background regions, we employ the prediction of the pretrained object detector~\cite{ren2024grounded}.
Also, the structure distance (SD) is employed to evaluate the structural difference between the source and translated images. 
It computes the Frobenius norm between the self-attention maps given by the DINO-ViT network output~\cite{caron2021emerging} using the source and translated images as inputs.

\subsection{Quantitative Results}
To compare the proposed method with state-of-the-art methods~\cite{hertz2023prompt, tumanyan2023plug, parmar2023zero}, we present quantitative results in \cref{tab:cmp_baselines}.
The table shows that our method consistently achieves the best performance in terms of BD and mostly outperforms the previous methods in terms of SD.  
As for CS, the proposed method shows the highest performance on the dog $\rightarrow$ cat task, while it ranks second in the remaining tasks.
Note that, because the CLIP similarity only reflects the fidelity to the target prompt without considering the similarity to the source images, it is not sufficiently discriminative to evaluate image-to-image translation performance by itself.
%We emphasize that it is important to consider a comprehensive comparison through multiple metrics. 
%We empirically observe that the slight differences in CLIP similarity do not lead to perceptible qualitative results. 
%despite achieving the best or the second-best performance for all cases,  
%the marginal difference imperceptable since the score depends largely on the generated prompts given by BLIP~\cite{li2022blip}.
%outperforms the previous algorithms in terms of BD and SD 
In addition, \cref{tab:addon_ptp,tab:addon_pnp,tab:addon_p2p} demonstrate that PIC is effective to improve the performance when incorporated into existing methods~\cite{hertz2023prompt, tumanyan2023plug, parmar2023zero}.

\subsection{Qualitative Results}
\cref{fig:comparisons} illustrates qualitative results generated by the proposed approach and other state-of-the-art methods~\cite{hertz2023prompt, tumanyan2023plug, parmar2023zero}. 
It presents that our method effectively preserves the background and structure of source images while selectively editing the region of interest. 
On the other hand, existing algorithms often fail to preserve the structure or background.
We present a failure case of our algorithm in the last row of \cref{fig:comparisons}, where the result from PIC is still favorable compared to others. 
\cref{fig:addon} demonstrates that PIC is effective to improve the previous methods when integrated into them.

\subsection{Inference Time}
To evaluate the inference time of each algorithm, we measure the wall-clock time using a single image on an NVIDIA A6000 GPU. 
As shown in \cref{tab:inference_time}, PIC is the most time-efficient even with its outstanding performance.

\begin{table}[t!]
	\centering
	\caption{
		Inference time comparisons between PIC and other state-of-the-art methods~\cite{hertz2023prompt, tumanyan2023plug, parmar2023zero}.%Inference Time (s)
	}
	\setlength\tabcolsep{16pt} 
	\scalebox{0.75}{
		%\resizebox{\linewidth}{!}{
			\begin{tabular}{c | c | c | c | c}
				%\begin{tabular}{l ccc ccc ccc ccc ccc}
				%\toprule 
				%\hline
				%& \multicolumn{3}{c}{DDIM}
				
				& PtP
				& PnP
				& P2P 
				& PIC (Ours) \\ 
				\hline
				
				Inference time (s)
				& 31.2 %& \bred{0.076} & 0.038 
				& \bred{24.4} % & 0.100 & \bred{0.032} 
				& 52.2 %& 0.127% & 0.099
				& \black{18.1} \\ %& \black{0.045} & \black{0.031} \\
				%& \black{0.293} & \black{0.045} & \black{0.031}\\
								
				\hline
				%\bottomrule 
			\end{tabular}
			
			%}
	}
	\label{tab:inference_time}
\end{table}

\subsection{Ablation Study}
\begin{table*}[t!]
	\centering
	\caption{
		Contribution of the noise correction and the prompt interpolation tested on LAION-5B dataset~\cite{schuhmann2022laion}.
		DDIM+PI synthesizes target images by replacing $\epth(\bxtgt_t, t, \bytgt)$ with $\epth(\bxtgt_t, t, \by_t)$ in the reverse DDIM process. 
		The model with the noise correction, DDIM+NC, substitutes $\epth(\bxtgt_t, t, \bytgt)$ for $\epth(\bxtgt_t, t, \by_t)$  without the consideration of the prompt interpolation.
		%the interpolated embedding
		%Ablation study results to compare the proposed method with Prompt-to-Prompt~\cite{hertz2023prompt}, Plug-and-Play~\cite{tumanyan2023plug}, and Pix2Pix-Zero~\cite{parmar2023zero} on real images sampled from the LAION-5B dataset~\cite{schuhmann2022laion} using the pretrained Stable Diffusion~\cite{rombach2022high}. 
		%and real images sampled from the LAION-5B dataset~\citep{schuhmann2022laion} for various tasks.
		%  text-driven image-to-image translation methods~\citep{hertz2022prompt, tumanyan2023plug, parmar2023zero} using the pre-trained Stable Diffusion~\citep{rombach2022high} and real images sampled from LAION 5B dataset~\citep{schuhmann2022laion} for various tasks. 
		%
		% For the the drawing $\rightarrow$ oil painting task, we do not report the BD score since the background can not be clearly defined.
		%and the lower value does not imply that the algorithm achieves good performance.
		%Black and red bold-faced numbers represent the best and second-best performances for each metric in each row. 
	}
	%\vspace{0.1in}
	%\vspace{-0.1in}
	\setlength\tabcolsep{2pt} 
	\scalebox{0.75}{
		%\resizebox{\linewidth}{!}{
			\begin{tabular}{l | ccc | ccc | ccc | ccc}
				%\begin{tabular}{l ccc ccc ccc ccc ccc}
				%\toprule 
				%\hline
				\multirow{2}{*}{Task} 
				%& \multicolumn{3}{c}{DDIM}
				& \multicolumn{3}{c|}{DDIM}
				& \multicolumn{3}{c|}{DDIM+PI}
				& \multicolumn{3}{c|}{DDIM+NC} 
				& \multicolumn{3}{c}{PIC (Ours)} \\ 
				%& \multicolumn{3}{c}{Avg} \\ %\\
				
				%\cline{2-4} \cline{5-7} \cline{8-10} \cline{11-13}
				%\cline{2-4} \cline{5-7} \cline{8-10} \cline{11-13} \cline{14-16}
				& CS ($\uparrow$)
				& BD ($\downarrow$)
				& SD ($\downarrow$)

				& CS ($\uparrow$)
				& BD ($\downarrow$)
				& SD ($\downarrow$)
				
				& CS ($\uparrow$)
				& BD ($\downarrow$)
				& SD ($\downarrow$)

				& CS ($\uparrow$)
				& BD ($\downarrow$)
				& SD ($\downarrow$)
				
				%& CS ($\uparrow$)
				%& BD ($\downarrow$)
				%& SD ($\downarrow$)
				
				\\ 
				\cline{1-13}
				%\cline{1-16}
				
				{dog $\rightarrow$ cat}
				% & ? & ? & ?
				& \bred{0.289} & 0.158 & 0.086 
				& \bred{0.289} & 0.130 & 0.070
				& \black{0.293} & \bred{0.054} & \bred{0.038}
				& \black{0.293} & \black{0.045} & \black{0.031} \\
				%& \black{0.293} & \black{0.045} & \black{0.031}\\

				{cat $\rightarrow$ dog}
				% & ? & ? & ?
				& 0.283 & 0.185 & 0.089
				& \bred{0.285} & 0.150 & 0.070
				& \black{0.288} & \bred{0.068} & \bred{0.041}
				& \black{0.288} & \black{0.057} & \black{0.033} \\
				%& \black{0.293} & \black{0.045} & \black{0.031}\\

				{horse $\rightarrow$ zebra}
				% & ? & ? & ?
				& 0.325	& 0.287	& 0.123
				& \bred{0.330}	& 0.214	& 0.097
				& \black{0.333}	& \bred{0.113}	& \bred{0.050}
				& 0.324	& \black{0.085}	& \black{0.037} \\
				%& \black{0.293} & \black{0.045} & \black{0.031}\\

				{zebra $\rightarrow$ horse}
				% & ? & ? & ?
				& \black{0.294}	& 0.295	& 0.104
				& \black{0.294}	& 0.254	& 0.097
				& \black{0.294}	& \bred{0.139}	& \bred{0.055}
				& \bred{0.292}	& \black{0.126}	& \black{0.050} \\
				%& \black{0.293} & \black{0.045} & \black{0.031}\\

				{tree $\rightarrow$ palm tree}
				& 0.304	& 0.234	& 0.088
				& 0.306 & \bred{0.222}	& 0.084
				& \bred{0.312} & \black{0.085}	& \bred{0.056}
				& \black{0.314}	& \black{0.085}	& \black{0.036} \\
				% & \black{0.293} & \black{0.045} & \black{0.031}\\
				
				{dog $\rightarrow$ dog w/glasses}
				& \black{0.318}	& 0.134	& 0.072
				& 0.310	& 0.132	& 0.065 
				& \bred{0.317}	& \bred{0.029}	& \bred{0.021}
				& 0.312	& \black{0.026}	& \black{0.016} \\
				%& \black{0.293} & \black{0.045} & \black{0.031} \\

				\hline
				{Average}
				& 0.302	& 0.216	& 0.094
				& 0.302	& 0.184	& 0.081 
				& \black{0.306}	& \bred{0.081}	& \bred{0.044} 
				& \bred{0.304}	& \black{0.071}	& \black{0.034} \\
				%& \black{0.293} & \black{0.045} & \black{0.031} \\
				
				\hline
				%\bottomrule 
			\end{tabular}
			
			%}
	}
	\label{tab:ablation}
\end{table*}

\subsubsection{Prompt Interpolation}
To analyze the impact of each component in our algorithm, we compare PIC with its three variations---DDIM, DDIM+PI, and DDIM+NC.
DDIM denotes a na\"ive application of the original DDIM algorithm~\cite{song2021denoising} to image-to-image translation.
DDIM+PI replaces the denoising network $\epth(\bxtgt_t, t, \bytgt)$ in Eq.~\eqref{eq:naive_target} with $\epth(\bxtgt_t, t, \by_t)$ using interpolated prompts $\by_t$ while DDIM+NC substitutes $\epth(\bxtgt_t, t, \bytgt)$ for $\epth(\bxtgt_t, t, \by_t)$ in Eq.~\eqref{eq:new_noise_prediction2} to compute the noise correction term without the proposed prompt interpolation.
As presented in \cref{tab:ablation}, DDIM+PI improves performance by using prompt interpolation compared with the standard DDIM and DDIM+NC is particularly helpful in preserving the background or structure of the source images by integrating the noise correction term.
Our algorithm, PIC, incorporating both the noise correction term and the prompt interpolation, achieves the best performance in the text-conditional image editing task. The qualitative results are presented in Fig. 6 of the appendix.

\subsubsection{Effect of Hyperparameter $\gamma$}
We study the effect of the hyperparameter $\gamma$ introduced in Eq.~\eqref{eq:new_noise_prediction2} to discuss the trade-off between the fidelity to the target prompt and the structure preservation.

\begin{figure*}[t!]
	\centering
	%\vspace{-5pt}
	%\vspace{0.2cm}
	\scalebox{1}{
	\includegraphics[width=\linewidth]{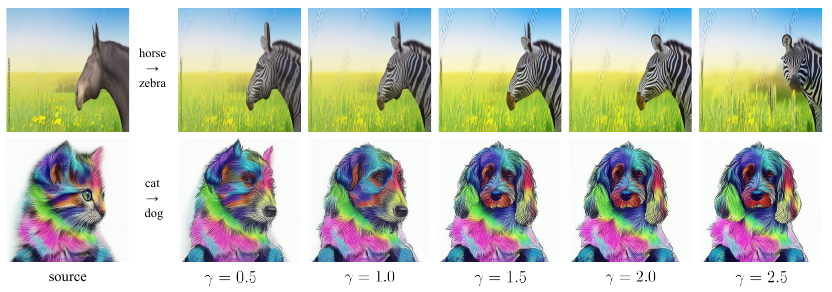}}
	%\vspace{-5pt}
	\caption{Qualitative results of the proposed method by varying $\gamma$ on data sampled from the LAION-5B dataset~\cite{schuhmann2022laion}, relying on the pretrained Stable Diffusion~\cite{rombach2022high}. 
		%LAION-5B dataset~\cite{schuhmann2022laion} using the pretrained Stable Diffusion~\cite{rombach2022high}.
		%Comparisons between various values of $\gamma$ in our proposed method. Optimal image quality usually observed at $\gamma=0.2$.
	}
	%Qualitative results of Pix2Pix on the Cityscapes dataset.
	% ``Output" denotes the ground-truth and ``Original" represents the uncompressed generator.}
%\vspace{-1mm}
\label{fig:gamma}
\end{figure*}

For the experiment related to $\gamma$, we explore five different values of $\gamma \in \{0.5, 1.0, 1.5, 2.0, 2.5 \}$ for PIC. \cref{fig:gamma} illustrates that our results are fairly consistent to the value of $\gamma$.
However, we observe that a low value of $\gamma$ tends to preserve the structure or background with relatively low fidelity, while a high value of $\gamma$ enhances fidelity at the expense of structure deformation. Note that we use $\gamma = 1.0$ throughout all experiments.

%
\iffalse
%
\begin{figure}
	\centering
	\includegraphics[width=1.0\linewidth]{./figures/ECCV_Rebuttal_SimpleLERP.pdf}
	\caption{Qualitative Comparison between PIC and PIC with simple LERP.}
	%Additional qualitative results of the proposed method and other text-driven image-to-image translation methods~\citep{hertz2022prompt, tumanyan2023plug, parmar2023zero} using the pre-trained Stable Diffusion~\citep{rombach2022high} and real images sampled from the LAION 5B dataset~\citep{schuhmann2022laion} on the cat $\rightarrow$ dog task.} 
\label{fig:simplelerp}
\end{figure}
%
\fi

% !TEX root = ./../main.tex
\section{Conclusion}
\label{sec:conclusion}
We presented a novel training-free approach for image-to-image translation based on text-to-image diffusion models.
We revised the original noise prediction network by incorporating a noise correction term with progressive interpolation of text embeddings.
Technically, the proposed noise prediction network for image-to-image translation consists of two parts: (a) the denoising network given the source latent and the source prompt and (b) a noise correction term defined as the difference between two noise predictions of the target latent conditioned on the progressively interpolated text embeddings and the source text embeddings. 
Extensive experiments demonstrate that the proposed algorithm achieves outstanding performance with reduced inference time and consistently improves existing techniques through the combination of those methods.

\section*{Acknowledgements}
This work was partly supported by LG AI Research, and by the National Research Foundation of Korea grant [No.2022R1A2C3012210] and the Institute for Information \& communications Technology Planning \& Evaluation (IITP) grants [RS-2022-II220959; RS-2021-II212068; RS-2021-II211343] funded by the Korea government (MSIT).

%\subsection{Ablation Study}

%\clearpage  % TODO REVIEW/FINAL: This \clearpage needs to be removed from both review and camera-ready versions.

\bibliographystyle{splncs04}
\bibliography{main}

\begin{thebibliography}{10}
\providecommand{\url}[1]{\texttt{#1}}
\providecommand{\urlprefix}{URL }
\providecommand{\doi}[1]{https://doi.org/#1}

\bibitem{brooks2023instructpix2pix}
Brooks, T., Holynski, A., Efros, A.A.: {Instructpix2pix: Learning to Follow
  Image Editing Instructions}. In: CVPR (2023)

\bibitem{caron2021emerging}
Caron, M., Touvron, H., Misra, I., J{\'e}gou, H., Mairal, J., Bojanowski, P.,
  Joulin, A.: {Emerging Properties in Self-Supervised Vision Transformers}. In:
  ICCV (2021)

\bibitem{esser2021taming}
Esser, P., Rombach, R., Ommer, B.: {Taming Transformers for High-Resolution
  Image Synthesis}. In: CVPR (2021)

\bibitem{gal2022stylegan}
Gal, R., Patashnik, O., Maron, H., Bermano, A.H., Chechik, G., Cohen-Or, D.:
  {StyleGAN-NADA: CLIP-Guided Domain Adaptation of Image Generators}. TOG
  (2022)

\bibitem{hertz2023prompt}
Hertz, A., Mokady, R., Tenenbaum, J., Aberman, K., Pritch, Y., Cohen-or, D.:
  {Prompt-to-Prompt Image Editing with Cross-Attention Control}. In: ICLR
  (2023)

\bibitem{hessel2021clipscore}
Hessel, J., Holtzman, A., Forbes, M., Le~Bras, R., Choi, Y.: {CLIPScore: A
  Reference-free Evaluation Metric for Image Captioning}. In: EMNLP (2021)

\bibitem{ho2020denoising}
Ho, J., Jain, A., Abbeel, P.: {Denoising Diffusion Probabilistic Models}. In:
  NeurIPS (2020)

\bibitem{ho2021classifier}
Ho, J., Salimans, T.: {Classifier-Free Diffusion Guidance}. In: NeurIPS 2021
  Workshop on Deep Generative Models and Downstream Applications (2021)

\bibitem{kawar2023imagic}
Kawar, B., Zada, S., Lang, O., Tov, O., Chang, H., Dekel, T., Mosseri, I.,
  Irani, M.: {Imagic: Text-based Real Image Editing with Diffusion Models}. In:
  CVPR (2023)

\bibitem{kim2022diffusionclip}
Kim, G., Kwon, T., Ye, J.C.: {DiffusionCLIP: Text-Guided Diffusion Models for
  Robust Image Manipulation}. In: CVPR (2022)

\bibitem{lee2023conditional}
Lee, H., Kang, M., Han, B.: {Conditional Score Guidance for Text-Driven
  Image-to-Image Translation}. In: NeurIPS (2023)

\bibitem{li2022blip}
Li, J., Li, D., Xiong, C., Hoi, S.: Blip: Bootstrapping language-image
  pre-training for unified vision-language understanding and generation. In:
  ICML (2022)

\bibitem{meng2022sdedit}
Meng, C., He, Y., Song, Y., Song, J., Wu, J., Zhu, J.Y., Ermon, S.: {SDEdit:
  Guided Image Synthesis and Editing with Stochastic Differential Equations}.
  In: ICLR (2022)

\bibitem{miyake2023negative}
Miyake, D., Iohara, A., Saito, Y., Tanaka, T.: {Negative-prompt Inversion: Fast
  Image Inversion for Editing with Text-guided Diffusion Models}. arXiv
  preprint arXiv:2305.16807  (2023)

\bibitem{mokady2023null}
Mokady, R., Hertz, A., Aberman, K., Pritch, Y., Cohen-Or, D.: {Null-text
  Inversion for Editing Real Images Using Guided Diffusion Models}. In: CVPR
  (2023)

\bibitem{parmar2023zero}
Parmar, G., Kumar~Singh, K., Zhang, R., Li, Y., Lu, J., Zhu, J.Y.: {Zero-Shot
  Image-to-Image Translation}. In: SIGGRAPH (2023)

\bibitem{radford2021learning}
Radford, A., Kim, J.W., Hallacy, C., Ramesh, A., Goh, G., Agarwal, S., Sastry,
  G., Askell, A., Mishkin, P., Clark, J., et~al.: {Learning Transferable Visual
  Models from Natural Language Supervision}. In: ICML (2021)

\bibitem{raffel2020exploring}
Raffel, C., Shazeer, N., Roberts, A., Lee, K., Narang, S., Matena, M., Zhou,
  Y., Li, W., Liu, P.J.: {Exploring the Limits of Transfer Learning with a
  Unified Text-to-Text Transformer}. JMLR  (2020)

\bibitem{aditya2022dalle2}
Ramesh, A., Dhariwal, P., Nichol, A., Chu, C., Chen, M.: Hierarchical
  text-conditional image generation with clip latents. arXiv preprint
  arXiv:2204.06125  (2022)

\bibitem{ren2024grounded}
Ren, T., Liu, S., Zeng, A., Lin, J., Li, K., Cao, H., Chen, J., Huang, X.,
  Chen, Y., Yan, F., et~al.: Grounded sam: Assembling open-world models for
  diverse visual tasks. arXiv preprint arXiv:2401.14159  (2024)

\bibitem{rombach2022high}
Rombach, R., Blattmann, A., Lorenz, D., Esser, P., Ommer, B.: {High-Resolution
  Image Synthesis with Latent Diffusion Models}. In: CVPR (2022)

\bibitem{ronneberger2015u}
Ronneberger, O., Fischer, P., Brox, T.: {U-Net: Convolutional Networks for
  Biomedical Image Segmentation}. In: MICCAI (2015)

\bibitem{saharia2022photorealistic}
Saharia, C., Chan, W., Saxena, S., Li, L., Whang, J., Denton, E.L.,
  Ghasemipour, K., Gontijo~Lopes, R., Karagol~Ayan, B., Salimans, T., et~al.:
  {Photorealistic Text-to-Image Diffusion Models with Deep Language
  Understanding}. In: NeurIPS (2022)

\bibitem{schuhmann2022laion}
Schuhmann, C., Beaumont, R., Vencu, R., Gordon, C., Wightman, R., Cherti, M.,
  Coombes, T., Katta, A., Mullis, C., Wortsman, M., et~al.: Laion-5b: An open
  large-scale dataset for training next generation image-text models. In:
  NeurIPS Datasets and Benchmarks Track (2022)

\bibitem{sohl2015deep}
Sohl-Dickstein, J., Weiss, E., Maheswaranathan, N., Ganguli, S.: {Deep
  Unsupervised Learning using Nonequilibrium Thermodynamics}. In: ICML (2015)

\bibitem{song2021denoising}
Song, J., Meng, C., Ermon, S.: {Denoising Diffusion Implicit Models}. In: ICLR
  (2021)

\bibitem{song2021score}
Song, Y., Sohl-Dickstein, J., Kingma, D.P., Kumar, A., Ermon, S., Poole, B.:
  {Score-Based Generative Modeling through Stochastic Differential Equations}.
  In: ICLR (2021)

\bibitem{tumanyan2022splicing}
Tumanyan, N., Bar-Tal, O., Bagon, S., Dekel, T.: {Splicing ViT Features for
  Semantic Appearance Transfer}. In: CVPR (2022)

\bibitem{tumanyan2023plug}
Tumanyan, N., Geyer, M., Bagon, S., Dekel, T.: {Plug-and-Play Diffusion
  Features for Text-Driven Image-to-Image Translation}. In: CVPR (2023)

\bibitem{van2017neural}
Van Den~Oord, A., Vinyals, O., et~al.: {Neural Discrete Representation
  Learning}. In: NIPS (2017)

\bibitem{zhang2018unreasonable}
Zhang, R., Isola, P., Efros, A.A., Shechtman, E., Wang, O.: {The Unreasonable
  Effectiveness of Deep Features as a Perceptual Metric}. In: CVPR (2018)

\end{thebibliography}
\end{document}